\def\year{2021}\relax
\documentclass[letterpaper]{article} 
\usepackage{aaai21}  
\usepackage{times}  
\usepackage{helvet} 
\usepackage{courier}  
\usepackage[hyphens]{url}  
\usepackage{graphicx} 
\usepackage{natbib}  
\usepackage{caption} 
\usepackage{microtype}
\usepackage{subfigure}
\usepackage{booktabs} 

\usepackage{amsmath,amsfonts,amsthm} 
\usepackage{algorithm}
\usepackage{algorithmic}
\usepackage{enumitem}
\usepackage{multirow}
\usepackage{bbm}

\usepackage[switch]{lineno}

\newtheorem{definition}{Definition}

\newtheorem{theorem}{Theorem}

\title{Beyond Class-Conditional Assumption: A Primary Attempt to Combat Instance-Dependent Label Noise}
\newcommand*\samethanks[1][\value{footnote}]{\footnotemark[#1]}
\author{
    Pengfei Chen,\textsuperscript{\rm 1}~
    Junjie Ye,\textsuperscript{\rm 2}\thanks{Corresponding authors.}~
    Guangyong Chen,\textsuperscript{\rm 3}\samethanks~
    Jingwei Zhao,\textsuperscript{\rm 2}~
    Pheng-Ann Heng\textsuperscript{\rm 1,3}\\
}
\affiliations {
    \textsuperscript{\rm 1} The Chinese University of Hong Kong\\
    \textsuperscript{\rm 2} VIVO AI Lab\\ 
    \textsuperscript{\rm 3} Shenzhen Key Laboratory of Virtual Reality and Human Interaction Technology, Shenzhen Institutes of Advanced Technology, Chinese Academy of Sciences\\
    \{pfchen, pheng\}@cse.cuhk.edu.hk, \{junjie.ye, jingwei.zhao\}@vivo.com, gy.chen@siat.ac.cn
}

\begin{document}
\maketitle

\begin{abstract}
Supervised learning under label noise has seen numerous advances recently, while existing theoretical findings and empirical results broadly build up on the class-conditional noise (CCN) assumption that the noise is independent of input features given the true label. In this work, we present a theoretical hypothesis testing and prove that noise in real-world dataset is unlikely to be CCN, which confirms that label noise should depend on the instance and justifies the urgent need to go beyond the CCN assumption.The theoretical results motivate us to study the more general and practical-relevant instance-dependent noise (IDN). To stimulate the development of theory and methodology on IDN, we formalize an algorithm to generate controllable IDN and present both theoretical and empirical evidence to show that IDN is semantically meaningful and challenging. As a primary attempt to combat IDN, we present a tiny algorithm termed self-evolution average label (SEAL), which not only stands out under IDN with various noise fractions, but also improves the generalization on real-world noise benchmark Clothing1M. Our code is released~\footnote{\url{https://github.com/chenpf1025/IDN}}. Notably, our theoretical analysis in Section 2 provides rigorous motivations for studying IDN, which is an important topic that deserves more research attention in future.
\end{abstract}

\section{Introduction}

Noisy labels are unavoidable in practical applications, where instances are usually labeled by workers on crowdsourcing platforms~\cite{yan2014learning,schroff2010harvesting}. Unfortunately, Deep Neural Networks (DNNs) can memorize noisy labels easily but generalize poorly on clean test data~\cite{zhang2017understanding}. Hence, how to mitigate the effect of noisy labels in the training of DNNs has attracted considerable attention recently. Most existing works, for their theoretical analysis or noise synthesizing in experiments, follow the class-conditional noise (CCN) assumption~\cite{scott2013classification,zhang2018generalized,menon2020can,ma2020normalized}, where the label noise is \textit{in}dependent of its input features conditional on the latent true label.

\begin{figure}[t]
	\begin{center}
		\centerline{\includegraphics[width=\columnwidth]{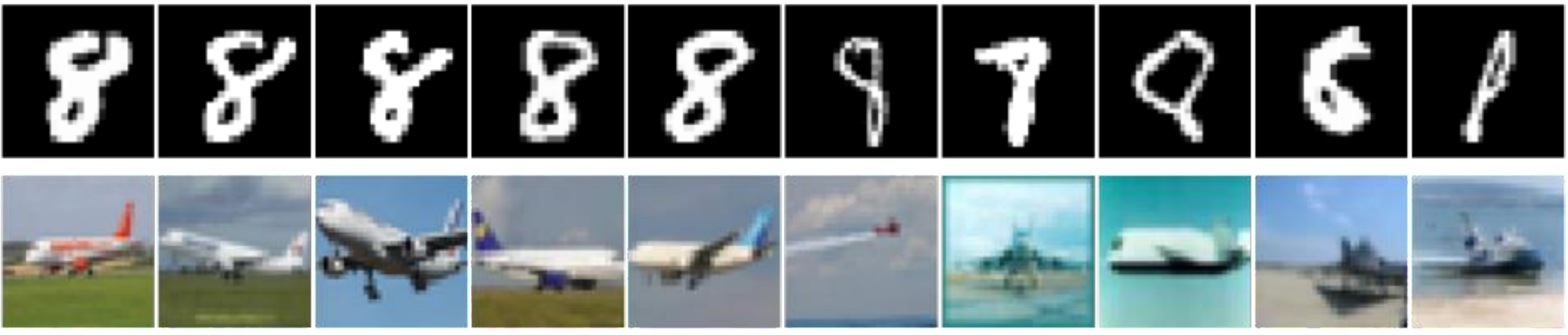}}
		\caption{Examples of \textit{8} in MNIST (first row) and \textit{Airplane} in CIFAR-10 (second row). It is problematic to assume a same probability of mislabeling for diverse samples in each class.}
		\label{fig_examples}
	\end{center}
	\vskip -0.3in
\end{figure}

In fact, instances with the same label can be entirely different, hence the probability of mislabeling should be highly dependent on the specific instance. As shown in the first row of Fig.~\ref{fig_examples}, the second right image is likely to be mislabeled as the number 6 and the fourth right image is likely to be manually mislabeled as the number 7; in the second row, the last image is more likely to be mislabeled as the ship.  In this paper, our \textit{first contribution} (Section~\ref{sec_ccn2idn}) is to present a theoretical hypothesis testing on the well-known real-world dataset, Clothing1M, to demonstrate the urgent need to go beyond the CCN assumption in practical applications. Meanwhile, we discuss the challenge of instance-dependent noise (IDN) with both theoretical and empirical evidence. Some pioneer efforts has been contributed to IDN, but most results are restricted to binary classification~\cite{menon2018learning,bootkrajang2018towards,cheng2020learning} or based on assumptions such as the noise is parts-dependent ~\cite{xia2020parts}.

To stimulate the development of theory and methodology on more practical-relevant IDN, we propose an algorithm to generate controllable IDN and present extensive characterizations of training under IDN, which is our \textit{second contribution} (Section~\ref{sec_idn}). Our \textit{third contribution} (Section~\ref{sec_seal}) is to propose an algorithm termed self-evolution average label (SEAL) to defend IDN, motivated by an experimental observation that the DNN's output corresponding to the latent true label can be activated with oscillation before memorizing noise. Specifically, SEAL provides instance-dependent label correction by averaging predictions of a DNN on each instance over the whole training process, then retrains a classifier using the averaged soft labels. The superior performance of SEAL is verified on extensive experiments, including synthetic/real-world datasets under IDN of different noise fractions, and the large benchmark Clothing1M~\cite{xiao2015learning} with real-world noise .

\section{From CCN to IDN - Theoretical Evidence}
\label{sec_ccn2idn}
\subsection{Preliminaries} 
Considering a $c$-class classification problem, let $\mathcal{X}$ be the feature space, $\mathcal{Y}=\{1,...,c\}$ be the label space, $(X,Y)\in\mathcal{X}\times\mathcal{Y}$ be the random variables with distribution $\mathcal{D}_{X,Y}$ and $D=\{(x_i,y_i)\}_{i=1}^{n}$ be a dataset containing i.i.d. samples drawn from $\mathcal{D}_{X,Y}$. In practical applications, the true label $Y$ may not be observable. Instead, we have an observable distribution of noisy labels $(X,\bar{Y})\sim\mathcal{\bar{D}}_{X,\bar{Y}}$ and a dataset $\bar{D}=\{(x_i,\bar{y}_i)\}_{i=1}^{n}$ drawn from it. A classifier $f:\mathcal{X}\rightarrow\mathbb{P}^c$ is defined by a DNN that outputs a probability distribution over all classes, where $\mathbb{P}^c=\{s\in\mathbb{R}_{+}^c:\|s\|_1=1\}$. By default, the probability is obtained by a softmax function~\cite{goodfellow2016deep} at the output of $f$.

\subsection{Beyond the CCN assumption}
The CCN assumption is commonly used in previous works, as clearly stated in theoretical analysis~\cite{blum1998combining,yan2017robust,patrini2016loss,zhang2018generalized,xu2019l_dmi,menon2020can,ma2020normalized} or inexplicitly used in experiments for synthetizing noisy labels ~\cite{han2018co,yu2019does,arazo2019unsupervised,li2020dividemix,lukasik2020does}.
Under the CCN assumption, the observed label $\bar{Y}$ is independent of $X$ conditioning on the latent true label $Y$.
\begin{definition}
    \label{def_ccn}
	(CCN Model) Under the CCN assumption, there is a noise transition matrix $M\in[0,1]^{c\times c}$ and we observe samples $(X,\bar{Y})\sim \mathcal{\bar{D}}=\mathrm{CCN}(\mathcal{D},M)$, where first we draw $(X,Y)\sim\mathcal{D}$ as usual, then flip $Y$ to produce $\bar{Y}$ according to the conditional probability defined by $M$, i.e., $\mathrm{Pr}(\bar{Y}=q|Y=p)=M_{p,q}$, where $p,q\in\mathcal{Y}$.
\end{definition}
We have seen various specific cases of CCN, including uniform/symmetric noise~\cite{ren2018learning,arazo2019unsupervised,chen2019meta,lukasik2020does}, pair/asymmetric noise~\cite{han2018co,chen2019understanding}, tri/column/block-diagonal noise~\cite{han2018masking}. Since the noise transition process is fully specified by a matrix $M$, one can mitigate the effect of CCN by modeling $M$~\cite{patrini2017making,hendrycks2018using,han2018masking,xia2019anchor,yao2019safeguarded}. Alternatively, several robust loss functions~\cite{natarajan2013learning,patrini2017making,zhang2018generalized,xu2019l_dmi} have been proposed and justified. Many other works do not focus on theoretical analysis, yet propose methods based on empirical findings or intuitions, such as sample selection~\cite{han2018co,song2019selfie,yu2019does}, sample weighting~\cite{ren2018learning} and label correction~\cite{ma2018dimensionality,arazo2019unsupervised}.

Intuitively, it can be problematic to assign the same noise transition probability to diverse samples in a same class, which is illustrated by examples in Fig.~\ref{fig_examples}. Theoretically, we can justify the need to go beyond the CCN assumption with the following theorem.

\begin{theorem} (CCN hypothesis testing)
	\label{theorem_ccn}
	Given a noisy dataset with n instances, considering random sampling a validation set $\bar{D}^{\prime}=\{(x_i,\bar{y}_i)\}_{i=1}^{m},\,m<n$, and training a network $f$ on the rest instances. After training, the validation error on $\bar{D}^{\prime}$ is $\hat{er}_{\bar{D}^{\prime}}^{0-1}[f]=\sum_{i=1}^{m}\frac{1}{m}\mathbbm{1}(f(x_i)\neq \bar{y}_i)$, where $\mathbbm{1}(\cdot)$ is the indicator function. Let $w_p=\mathrm{Pr}[Y=p],\,p\in\mathcal{Y}$ be the fraction of samples in each class, then the following holds given the CCN assumption.
	\begin{equation}
	\label{eq_ccn}
	    \mathrm{Pr}\left[1-\sum_{p=1}^{c}w_p\max_{q\in\mathcal{Y}}M_{p,q}-\hat{er}_{\bar{D}^{\prime}}^{0-1}[f]\geq\varepsilon\right]\leq e^{-2m\varepsilon^2}
	\end{equation}
	\begin{proof}
	Let $er_{\mathcal{\bar{D}}}^{0-1}[f]=\mathbb{E}_{(X,\bar{Y})\sim\mathcal{\bar{D}}}\mathbbm{1}(f(X)\neq\bar{Y})$ be the generalization error on noisy distribution. For any $f$, the CCN assumption implies $f(X)$ is independent of $\bar{Y}$ conditional on $Y$, then we have,
	\begin{equation}
	\nonumber
	\begin{aligned}
	    er_{\mathcal{\bar{D}}}^{0-1}[f] &= 1-\mathbb{E}_{(X,\bar{Y})\sim\mathcal{\bar{D}}}\mathbbm{1}(f(X)=\bar{Y})\\
	    &=1-\sum_{p=1}^{c}w_p\mathrm{Pr}[f(X)=\bar{Y}|Y=p]\\
	    &=1-\sum_{p=1}^{c}w_p\sum_{q=1}^c\mathrm{Pr}[f(X)=q,\bar{Y}=q|Y=p]\\
	    &=1-\sum_{p=1}^{c}w_p\sum_{q=1}^c\mathrm{Pr}[f(X)=q|Y=p]\cdot M_{p,q}\\
	    &\geq 1-\sum_{p=1}^{c}w_p\max_{q\in\mathcal{Y}}M_{p,q}.\\
	\end{aligned}
	\end{equation}
	Note that the error $\hat{er}_{\bar{D}^{\prime}}^{0-1}[f]$ is estimated on validation samples that are not used when training $f$, hence $\mathbbm{1}(f(x_i)\neq \bar{y}_i),i=1,\cdots,m$, are $m$ i.i.d Bernoulli random variables with expectation $er_{\mathcal{\bar{D}}}^{0-1}[f]$. Using Hoeffding’s inequality, we have, $\forall\varepsilon>0$,
	\begin{equation}
	\nonumber
	\begin{aligned}
	&\mathrm{Pr}\left[1-\sum_{p=1}^{c}w_p\max_{q\in\mathcal{Y}}M_{p,q}-\hat{er}_{\bar{D}^{\prime}}^{0-1}[f]\geq\varepsilon\right]\\
	&\leq \mathrm{Pr}\left[er_{\mathcal{\bar{D}}}^{0-1}[f]-\hat{er}_{\bar{D}^{\prime}}^{0-1}[f]\geq\varepsilon\right]\leq e^{-2m\varepsilon^2}.
	\end{aligned}
	\end{equation}
	\end{proof}
\end{theorem}

Now we apply Theorem~\ref{theorem_ccn} to the widely used noise benchmark Clothing1M, which contains $1M$ noisy training samples of clothing images in 14 classes. We train a ResNet-50 on $500K$ random sampled instances, validate on the rest and obtain a validation accuracy $\hat{er}_{\bar{D}^{\prime}}^{0-1}[f]=0.1605$. The original paper~\cite{xiao2015learning} provides additional refined labels and a noise confusion matrix that can be an estimator of $M$ under the CCN assumption. Moreover, we estimate
$w_p$ using the proportion of labels on the $14k$ refined subset. In this way, we get $1-\sum_{p=1}^{c}w_p\max_{q\in\mathcal{Y}}M_{p,q}=0.3817$. Now we have $1-\sum_{p=1}^{c}w_p\max_{q\in\mathcal{Y}}M_{p,q}-\hat{er}_{\bar{D}^{\prime}}^{0-1}[f]=0.2212$. By substituting $\varepsilon=0.2212$ to Eq.~(\ref{eq_ccn}), we see that this result happens with {\bf probability lower than $\mathbf{10^{-21250}}$}, which is statistically impossible. This contradiction implies that the CCN assumption does not hold on Clothing1M. In fact, this result is explainable by analyzing the difference between CCN and IDN. Under the CCN assumption, in each class, label noise is independent of input features, hence the network can not generalize well on such independent noise. Thus we derive a high validation error conditioning on CCN. While practically, we obtain an error much lower than the derived one, which implies that the network learned feature-dependent noise that can be generalized to the noisy validation set.

\begin{figure}[t]
	\begin{center}
		\centerline{\includegraphics[width=0.6\columnwidth]{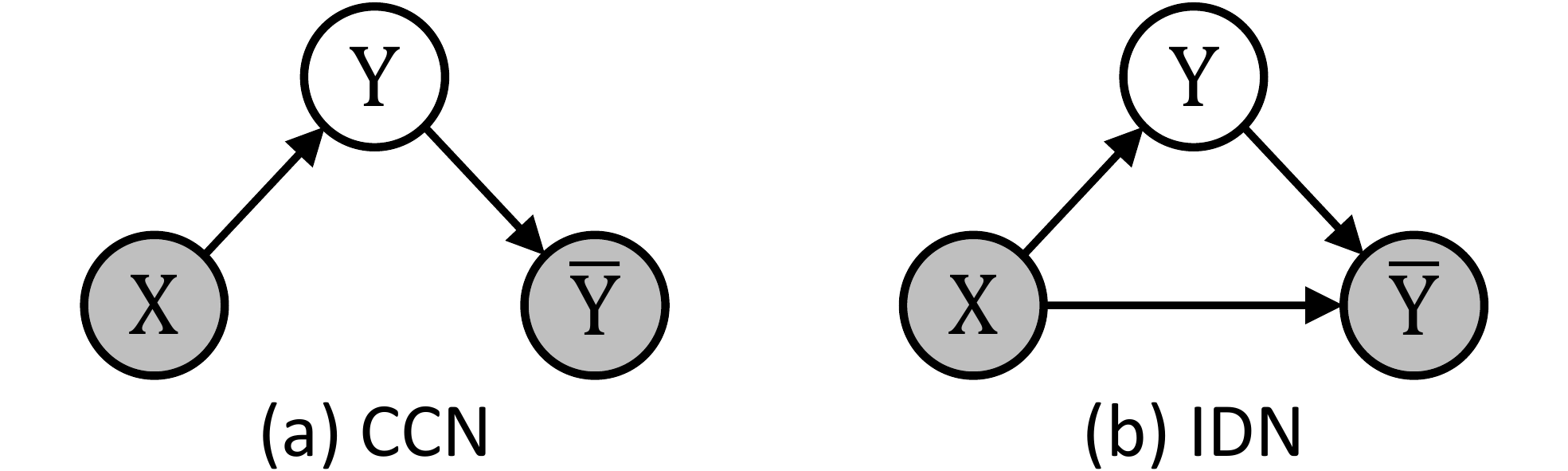}}
		\vskip -0.1in
		\caption{The graphical model of label noise.}
		\label{fig_graphical}
	\end{center}
	\vskip -0.2in
\end{figure}

\subsection{The IDN model and its challenges}
Now both theoretical evidence and the intuition imply that label noise should be dependent on input features, yet limited research efforts have been devoted to IDN. For binary classification under IDN, there have been several pioneer theoretical analysis on robustness~\cite{menon2018learning,bootkrajang2018towards} and sample selection methods~\cite{cheng2020learning}, mostly restricted to small-scale machine learning such as logistic regression. Under deep learning scenario, \citet{xia2020parts} combat IDN by assuming that the noise is parts-dependent, hence they can estimate the noise transition for each part. \citet{thulasidasan2019combating} investigate the co-occurrence of noisy labels with underlying features by adding synthetic features, such as the smudge, to mislabeled instances. However, this is not the typical realistic IDN where the noisy label should be dependent on inherent input features. 

As presented in Definition~\ref{def_idn}, we can model instance-dependent mislabelling among given classes, where the noise transition matrix is a function of $X$. Note that both IDN and CCN consider close-set noise, as contrast to a specific label noise termed open-set noise~\cite{wang2018iterative}, where the noisy instances does not belong to any considered classes. The graphical model of label noise is shown in Fig.~\ref{fig_graphical}. CCN can be seen as a degenerated case of IDN such that all instances have the same noise transition matrix. 
\begin{definition}
    \label{def_idn}
	(IDN Model) Under the IDN model, $M:\mathcal{X}\rightarrow[0,1]^{c\times c}$ is a function of $X$. We observe samples $(X,\bar{Y})\sim \mathcal{\bar{D}}=\mathrm{IDN}(\mathcal{D},M)$, where first we draw $(X,Y)\sim\mathcal{D}$ as usual, then flip $Y$ to produce $\bar{Y}$ according to the conditional probability defined by $M(X)$ , i.e., $\mathrm{Pr}(\bar{Y}=q|Y=p)=M_{p,q}(X)$, where $p,q\in\mathcal{Y}$.
\end{definition}

Many existing robust loss functions~\cite{natarajan2013learning,patrini2017making,zhang2018generalized,xu2019l_dmi} have theoretical guarantees derived from the CCN assumption but not IDN. Some sample selection algorithms~\cite{malach2017decoupling,han2018co,yu2019does,li2020dividemix}, targeting at selecting clean samples from the noisy training set, work quite well under CCN. Though these methods does not directly rely on the CCN assumption, it can be more challenging to identify clean samples under IDN since the label noise is correlated with inherent input features that result in confusion. 

Theoretically, we  show that
\textit{the optimal sample selection exists under CCN but may fail under IDN.}
This is because under IDN, even if we select all clean samples, the distribution of $X$ can be different to its original distribution. While for CCN, we can select an optimal subset in theory. The key issue is whether the following holds for any $p\in\mathcal{Y}$.
\begin{equation}
    \label{eq_supp}
    supp(P(X|\bar{Y}=Y, Y=p))\overset{?}{=}supp(P(X|Y=p)),
\end{equation}
where $supp(\cdot)$ is the support of a distribution. For CCN, since $X$ is independent of $\bar{Y}$ conditioning $Y$, the equality in Eq.~(\ref{eq_supp}) holds. While for IDN, it mostly does not hold. For example, if samples near the decision boundary are more likely to be mislabeled, then the support is misaligned for clean samples, which means learning with selected clean samples is statistically inconsistent~\cite{cheng2020learning}. More characterizations will be presented in the next section.

\section{A typical controllable IDN}
\label{sec_idn}

\subsection{Enabling controllable experiments}
The rapid advance of research on CCN not only attributes to simplicity of the noise model, but also the easy generation process of synthetic noise. We are able to conduct experiments on synthetic CCN with varying noise fractions by randomly flipping labels according to the conditional probability defined by $M$, which enable us to characterize DNNs trained with CCN~\cite{arpit2017closer,chen2019understanding}, develop algorithms accordingly and quickly verify the idea. Similarly, it is desired to easily generate IDN with any noise fraction for any given benchmark dataset. A practical solution is to model IDN using DNNs' prediction error because the error is expected to be challenging for DNNs. To yield calibrated softmax output for IDN generation, \citet{berthon2020confidence} train a classifier on a small subset, calibrate the classifier on another clean validation set~\cite{guo2017calibration}, and then use predictions on the rest instances to obtain noisy labels. It does not generate noise for the whole dataset and the noise largely depends on the small training subset.

To stimulate the development of theory and methodology, we propose a novel IDN generator in Algorithm~\ref{alg_gen_idn}. Our labeler follows the intuition that `hard' instances are more likely to be mislabeled~\cite{du2015modelling,menon2018learning}. Given a dataset $D=\{(x_i,y_i)\}_{i=1}^{n}$ with labels believed to be clean, we normally train a DNN for $T$ epochs and get a sequence of networks with various classification performance. For each instance, if many networks predict a high probability on a class different to the labeled one, it means that it is hard to clearly distinguish the instance from this class. Therefore, we can compute the score of mislabeling $N(x)$ and the potential noisy label $\tilde{y}(x)$ as follow:
\begin{equation}
\label{eq_gen}
\begin{split}
S&=\sum_{t=1}^{T}S^{t}/T\in\mathbb{R}^{n\times c},\\
N(x_i)&=\max_{k\neq y_i}S_{i,k},\quad\tilde{y}(x_i)=\arg\max_{k\neq y_i}S_{i,k}, 
\end{split}
\end{equation}
where $S^{t}=[f^{t}(x_i)]_{i=1}^{n}$ is DNN's output at $t$-th epoch. The average prediction here reveals the DNN's confusion on instances throughout training. We flip the label of $p\%$ instances with highest mislabeling scores, where $p$ is the targeted noise fraction. In essence, Algorithm~\ref{alg_gen_idn} uses predictions of the DNN to synthetize noisy labels, while it stands out for being able to generate noisy labels of any noise ratio for the whole training set, requiring simply a single round of training on given labels. The noise is instance-dependent since it comes from the prediction error on each instance. Moreover, it is a typical challenging IDN since the error is exactly the class hard for the DNN to distinguish. In Appendix~\ref{app_eg_idn}, examples of noisy samples show that the noise is semantically meaningful.

\begin{algorithm}[t]
	\caption{IDN Generation.}
	\label{alg_gen_idn}
	\begin{algorithmic}
		\STATE \textbf{Input:} Clean samples $D=\{(x_i,y_i)\}_{i=1}^{n}$, a targeted noise fraction $p$, epochs $T$.
		\STATE Initialize a network $f$.
		\FOR{$t=1$ \textbf{to} $T$}
		\FOR{batches $\{(x_i,y_i)\}_{i\in\mathcal{B}}$}
		\STATE Train $f$ on $\{(x_i,y_i)\}_{i\in\mathcal{B}}$ using cross-entropy loss:
		\centering{$\mathcal{L}_{CE}=-\frac{1}{\lvert\mathcal{B}\rvert}\sum_{i\in\mathcal{B}}\log(f_{y_i}^{t}(x_i))$}
		\ENDFOR
		\STATE Record output $S^{t}=[f^{t}(x_i)]_{i=1}^{n}\in\mathbb{R}^{n\times c}$.
		\ENDFOR
		\STATE Compute $N(x_i)$, $\tilde{y}(x_i)$ using $\{S^{t}\}_{t=1}^{T}$ (Eq.~(\ref{eq_gen})).
		\STATE Compute the index set $\mathcal{I} = \{p\%\arg\max_{1\leq i\leq n}N(x_i)\}$.
		\STATE Flip $\bar{y}_i=\tilde{y}_i$ if $i\in\mathcal{I}$ else keep $\bar{y}_i=y_i$.
		\STATE \textbf{Output:} A dataset with IDN: $\bar{D}=\{(x_i,\bar{y}_i)\}_{i=1}^{n}$.
	\end{algorithmic}
\end{algorithm}

\subsection{Characterizations of training with IDN}
\label{sec_idn_charater}
To combat label noise, we can firstly characterize behaviors of DNNs trained with noise. For example, the memorization effect~\cite{arpit2017closer} under CCN claims that DNNs tend to learn simple and general patterns first before memorizing noise, which has motivated extensive robust training algorithms. While our understanding on IDN is still limited. Here we present some empirical findings on IDN, to help researchers understand the behaviors of DNNs trained with IDN and to motivate robust training methods. We conduct experiments on MNIST and CIFAR-10 under IDN with varying noise fractions generated by Algorithm~\ref{alg_gen_idn}. For CCN, we use the most studied uniform noise. In all experiments throughout this paper, the DNN model and training hyperparameters we use are consistent. More details on experimental settings are summarized in Section~\ref{sec_exp_setup} and Appendix~\ref{app_exp_setup}.

\begin{figure*}[t]
	\begin{center}
		\centerline{\includegraphics[width=1.6\columnwidth]{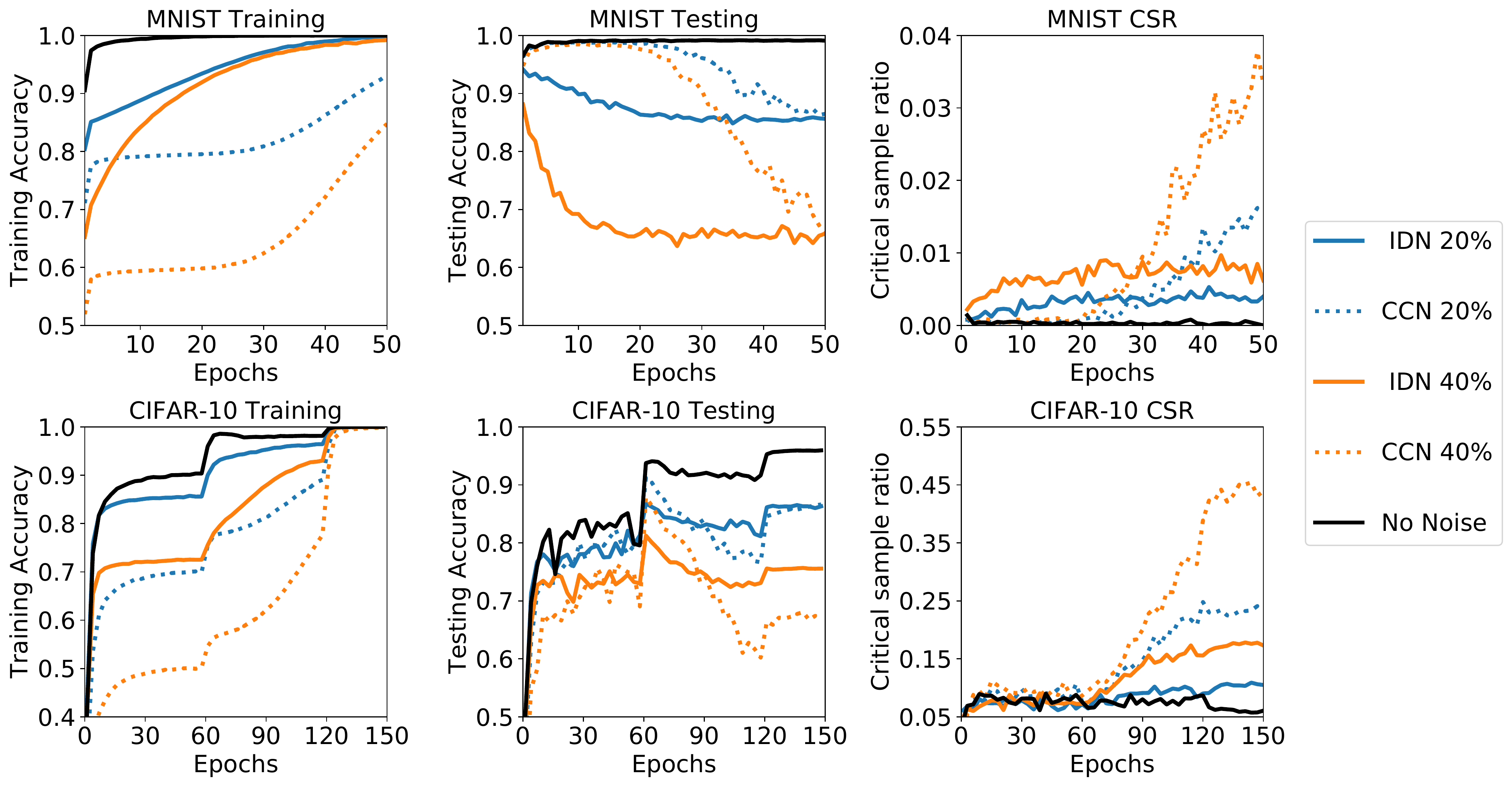}}
		\vskip -0.15in
		\caption{Training/testing accuracy and critical sample ratio throughout training on IDN and CCN with varying noise fractions.
		}
		\label{fig_train_curve}
	\end{center}
	\vskip -0.3in
\end{figure*}

\paragraph{It is easier for DNNs to fit IDN.}
Firstly, let us focus on the training/testing curves in Fig.~\ref{fig_train_curve}. For IDN and CCN with the same noise fraction, the training accuracy is higher under IDN. This implies that it is easier for DNNs to fit IDN. The finding is consistent with our intuition since noisy labels under IDN are highly correlated with input features that can mislead DNNs.  In this sense, IDN is more difficult to mitigate because the feature-dependent noise is very confusing for DNNs, which can easily result in overfitting. Moreover, the peak testing accuracy before convergence, which implies the DNN learns general patterns first~\cite{arpit2017closer}, is much lower under IDN. This suggests that due to DNNs can fit IDN easily, the generalization performance degenerates at early stages of training. The observation is consistent with the findings on real-world noise presented by~\citet{jiang2020beyond}.

\paragraph{The memorization effect is less significant.}
The memorization effect~\cite{arpit2017closer} is a critical phenomenon of DNNs trained with CCN: DNNs first learn \textit{simple} and \textit{general} patterns of the real data before fitting noise. It has motivated extensive robust training algorithms. The memorization effect is characterized by the testing accuracy and critical sample ratio (CSR)~\cite{arpit2017closer} during training, where CSR estimates the density of decision boundaries. A sample $x$ is a critical sample if there exists a $\hat{x}$, s.t.,
\begin{equation}
\arg\max_k f_k(x) \neq \arg\max_k f_k(\hat{x}),\,\mathrm{s.t.}\,\| x - \hat{x}\|_{\infty}\leq r.
\end{equation}
The curves of testing accuracy and CSR presented in Fig.~\ref{fig_train_curve} show typical characterizations of the memorization effect. Similar to CCN, the model achieves maximum testing accuracy before memorizing all training samples under IDN, which suggests that DNNs can learn general patterns first. Moreover, the CSR increases during training, suggesting that DNNs learn gradually more complex hypotheses. It is worth noting that under IDN, both peak testing accuracy and CSR are lower, and the gap between peak and converged testing accuracy is smaller. On MNIST, the testing accuracy decreases since very early stage of training, suggesting that the memorizing of noise dominates learning of real data. Therefore, we conclude that the memorization effect still exists under IDN, but it is less significant compared to CCN.

\begin{figure}[t]
	\begin{center}
		\centerline{\includegraphics[width=\columnwidth]{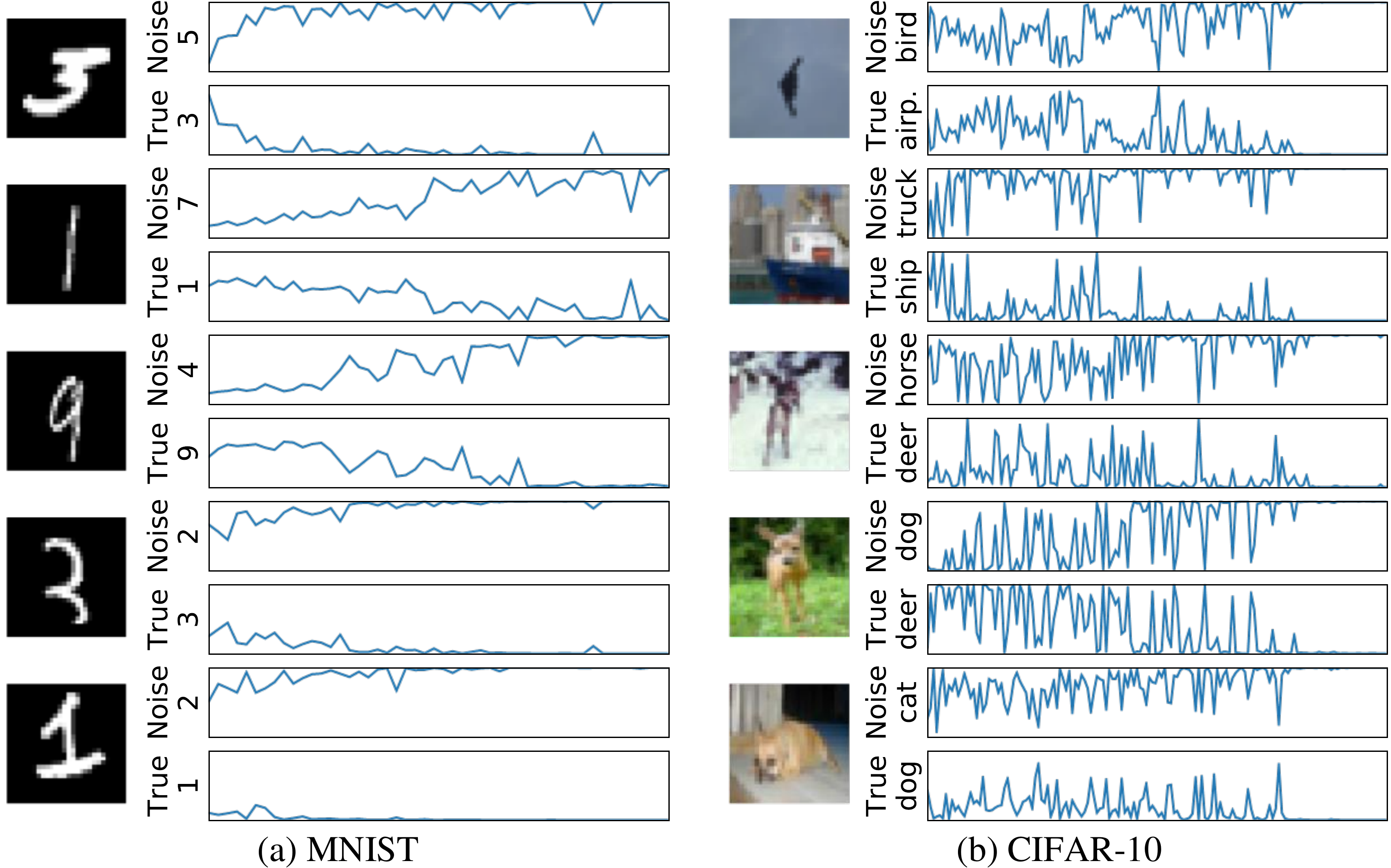}}
		\vskip -0.1in
		\caption{Examples of softmax outputs on the noisy label and latent true label. The x-axis is training epoch and the y-axis is DNN's output probability. The airp. is airplane for short.}
		\label{fig_train_examples}
	\end{center}
	\vskip -0.3in
\end{figure}

\paragraph{Individual study: instance-level memorization.}
Apart from showing the memorization effect for the whole training set, we are interested in how memorization happens for individual instances. As an individual study, we train DNNs under $20\%$ IDN and show examples in Fig.~\ref{fig_train_examples}. We plot the entry of softmax output corresponding to the noisy label and true label throughout training. DNNs will eventually memorize the wrong label, while during training, the output corresponding to the true label can be largely activated with oscillation. The intensity of oscillation and the epoch when the memorization happens is quite different for each instance.

\section{SEAL: a primary attempt to combat IDN}
\label{sec_seal}

\subsection{Methods}
\begin{algorithm}[tb]
	\caption{An iteration of SEAL.}
	\label{alg_seal}
	\begin{algorithmic}
		\STATE \textbf{Input:} Noisy samples $\bar{D}=\{(x_i,\bar{y}_i)\}_{i=1}^{n}$, epochs $T$, soft labels from the last iteration $\bar{S}$ (optional).
		\STATE Initialize a network $f$.
		\IF{$\bar{S}$ is not available}
		\STATE $\#\,$\textit{The initial iteration, using one-hot noisy labels} 
		$\bar{S}=[e_{\bar{y}_i}]_{i=1}^{n}\in\mathbb{R}^{n\times c}$ where $e_{\bar{y}_i}$ \textit{is the one-hot label}.
		\ENDIF
		\FOR{$t=1$ \textbf{to} $T$}
		\FOR{batches $\{(x_i,\bar{S}_i)\}_{i\in\mathcal{B}}$}
		\STATE Train $f$ on $\{(x_i,\bar{S}_i)\}_{i\in\mathcal{B}}$ using the loss:\\
		\centering{$\mathcal{L}_{SEAL}=-\frac{1}{\lvert\mathcal{B}\rvert}\sum_{i\in\mathcal{B}}\sum_{k=1}^{c}\bar{S}_{i,k}\log(f_{k}^{t}(x_i))$}
		\ENDFOR
		\STATE Record output $\bar{S}^{t}=[f^{t}(x_i)]_{i=1}^{n}\in\mathbb{R}^{n\times c}$.
		\ENDFOR
		\STATE Update $\bar{S}=\sum_{t=1}^{T}\bar{S}^{t}/T\in\mathbb{R}^{n\times c}$.
		\STATE \textbf{Output:} Trained $f$, $\bar{S}$ (can be used in next iteration).
	\end{algorithmic}
\end{algorithm}

To mitigate the effect of label noise, we propose a practical algorithm termed self-evolution average label (SEAL). SEAL provides instance-dependent label correction by averaging predictions of a DNN on each instance over the whole training process, then retrains a classifier using the averaged soft labels. An iteration of SEAL is outlined in Algorithm~\ref{alg_seal}, and we can apply SEAL with multiple iterations.

Here we discuss the intuitions of SEAL. Without loss of generality, assume there exists a latent optimal distribution of true label for each instance. Let $S_{i}^{*}\in\mathbb{P}^c$ be the latent optimal label distribution of the $i$-th instance. $S_{i}^{*}$ can be one-hot for a confident instance and be soft otherwise. Intuitively, we can image $S_{i}^{*}$ as the output of an oracle DNN. Considering training a DNN on a $c$-class noisy dataset $\bar{D}=\{(x_i,\bar{y}_i)\}_{i=1}^{n}$ for sufficient many $T$ epochs until converged, we let $f^{t}(x_i)$ be the output on $x_i$ at $t$-th epoch. Based on the oscillations show in Figure~\ref{fig_train_examples}, we roughly approximate the output on $x_i$ at $t$-th epoch as
\begin{equation}
\label{eq_pred}
f^{t}(x_i) = \alpha_{i}^{t}\omega_{i}^{t}+ (1-\alpha_{i}^{t})e_{\bar{y}_i}, 
\end{equation}
where $t\in\{1,2,\cdots,T\}$, $e_{\bar{y}_i}$ is the one-hot label, $\alpha_{i}^{t}\in[0,1]$ are coefficients dependent on instances and the network, $w_{i}^{t}\in\mathbb{P}^c$ are i.i.d. random vectors with $\mathbb{E}[w_{i}^{t}]=S_{i}^{*}$. The approximation may be inaccurate at the early stage of training because the network does not learn useful features and it is better to add a term for random predictions in the approximation. Still, we ignore this term because random predictions do not introduce bias toward any class and the effect is mitigated by taking the average. With the approximation, we intuitively compare with the prediction of a random epoch $f^{\tau}(x_i)$, where $\tau$ is a random epoch such that $\mathrm{Pr}[\tau=t]=1/T,\forall t\in\{1,2,\cdots,T\}$. Let $\bar{S}\in\mathbb{R}^{n\times c}$ be the soft labels obtained by SEAL and $\|\cdot\|$ denote a norm on $\mathbb{P}^c$, it is not difficult to see that for any training instance $x_i$,
\begin{equation}
\label{eq_seal0_b}
\|\mathbb{E}[\bar{S}_{i}]-S_{i}^{*}\|\leq\|e_{\bar{y}_i}-S_{i}^{*}\|,\\ 
\end{equation}
\vskip -0.2in
\begin{equation}
\label{eq_seal0_c}
\mathrm{var}(\bar{S}_{i,k})\leq\mathrm{var}(f_{k}^{\tau}(x_i)),\,\forall k\in\{1,2,\cdots,c\}.
\end{equation}
That is, SEAL yields instance-dependent label correction that is expected to be better than the given noisy labels and the label correction has lower variance due to taking the average.

We can run SEAL for multiple iteration to further correct the noise, and we term this `self-evolution'. We take the soft label (denoted as $\bar{S}_{i}^{[m]},m\geq0$) of the last iteration as input and output $\bar{S}_{i}^{[m+1]}$. Using similar approximation as Eq.~(\ref{eq_pred}) by replacing the training label $e_{\bar{y}_i}$ with $\bar{S}_{i}^{[m]}$, SEAL is expected to produce labels that gradually approach the optimal ones,
\begin{equation}
\|\mathbb{E}[\bar{S}_{i}^{[m+1]}]-S_{i}^{*}\| \leq \|\bar{S}_{i}^{[m]}-S_{i}^{*}\|.
\end{equation}
A concern of SEAL is the increased computational cost due to retraining the network. In experiments, we focus on verifying the idea of SEAL and we retrain networks from the scratch in each iteration to show the evolution under exactly the same training process, resulting in scaled computational cost. While in practice, we may save computational cost by reserving the best model (e.g., using a noisy validation set) and training for less epochs.

\subsection{SEAL v.s. related methods}
\label{sec_seal_advantage}
Using predictions of DNNs has long been adopted in distillation~\cite{hinton2015distilling} and robust training algorithms that use pseudo labels~\cite{reed2015training,ma2018dimensionality,tanaka2018joint,song2019selfie,nguyen2019self,arazo2019unsupervised}. SEAL provides an elegant solution that is simple, effective and has empirical and theoretical intuitions. Taking the average of predictions, motivated by the activation and oscillation of softmax output at the entry of true label, provides label correction. SEAL is different to vanilla distillation~\cite{hinton2015distilling}: in the presence of label noise, simply distilling knowledge from a converged teacher network, which memorizes noisy labels, can not correct the noise.

Compared with existing pseudo-labeling methods, SEAL does not require carefully tuning hyperparameters to ensure that (i) the DNN learns enough useful features and (ii) the DNN dose not fit too much noise. It is challenging to compromise between (i) and (ii) in learning with IDN. We have shown that the memorization on correct/noisy labels can be quite different for each training instance. However, the above (i) and (ii) are typically required in existing methods~\cite{reed2015training,ma2018dimensionality,tanaka2018joint,song2019selfie,nguyen2019self,arazo2019unsupervised}. For example, one usually needs to tune a warm-up epoch~\cite{tanaka2018joint,song2019selfie,arazo2019unsupervised} before which no label correction is applied. A small warm-up epoch results in underfitting on useful features while a large one yields overfitting on noise. Worse still, one may need to tune an adaptive weight during training to determine how much we trust predictions of the DNN~\cite{reed2015training,ma2018dimensionality}. As theoretically shown by~\citet{dong2019distillation}, the conditions are very strict for DNNs to converge and not to fit noise.

When implementing SEAL, there is no specific hyperparameters other than the canonical hyperparameters such as the training epoch and learning rate. To determine these canonical hyperparameters, we simply need to examine the training accuracy on the noisy dataset. Since SEAL averages predictions throughout training, the label correction can be effective even if the DNN memorizes noise when converged. Therefore, our criterion of choosing hyperparameters is to make sure the training accuracy is converged and it is as high as possible. Moreover, the model architecture and training hyperparameters can be shared in each iteration of SEAL.

\subsection{Empirical evaluation}
\label{sec_seal_exp}

\paragraph{Experimental setup.}
\label{sec_exp_setup}
Our experiments focus on challenging IDN and real-world noise. We demonstrate the performance of SEAL on MNIST and CIFAR-10~\cite{krizhevsky2009learning} with varying IDN fractions as well as large-scale real-world noise benchmark Clothing1M~\cite{xiao2015learning}. We use a CNN on MNIST and the Wide ResNet 28$\times$10~\cite{zagoruyko2016wide} on CIFAR-10. On Clothing1M, we use the ResNet-50~\cite{he2016deep} following the benchmark setting~\cite{patrini2017making,tanaka2018joint,xu2019l_dmi}. More details on experiments can be found in Appendix~\ref{app_exp_setup}.

\paragraph{SEAL corrects label noise.}
\begin{figure}[!t]
	\begin{center}
		\centerline{\includegraphics[width=\columnwidth]{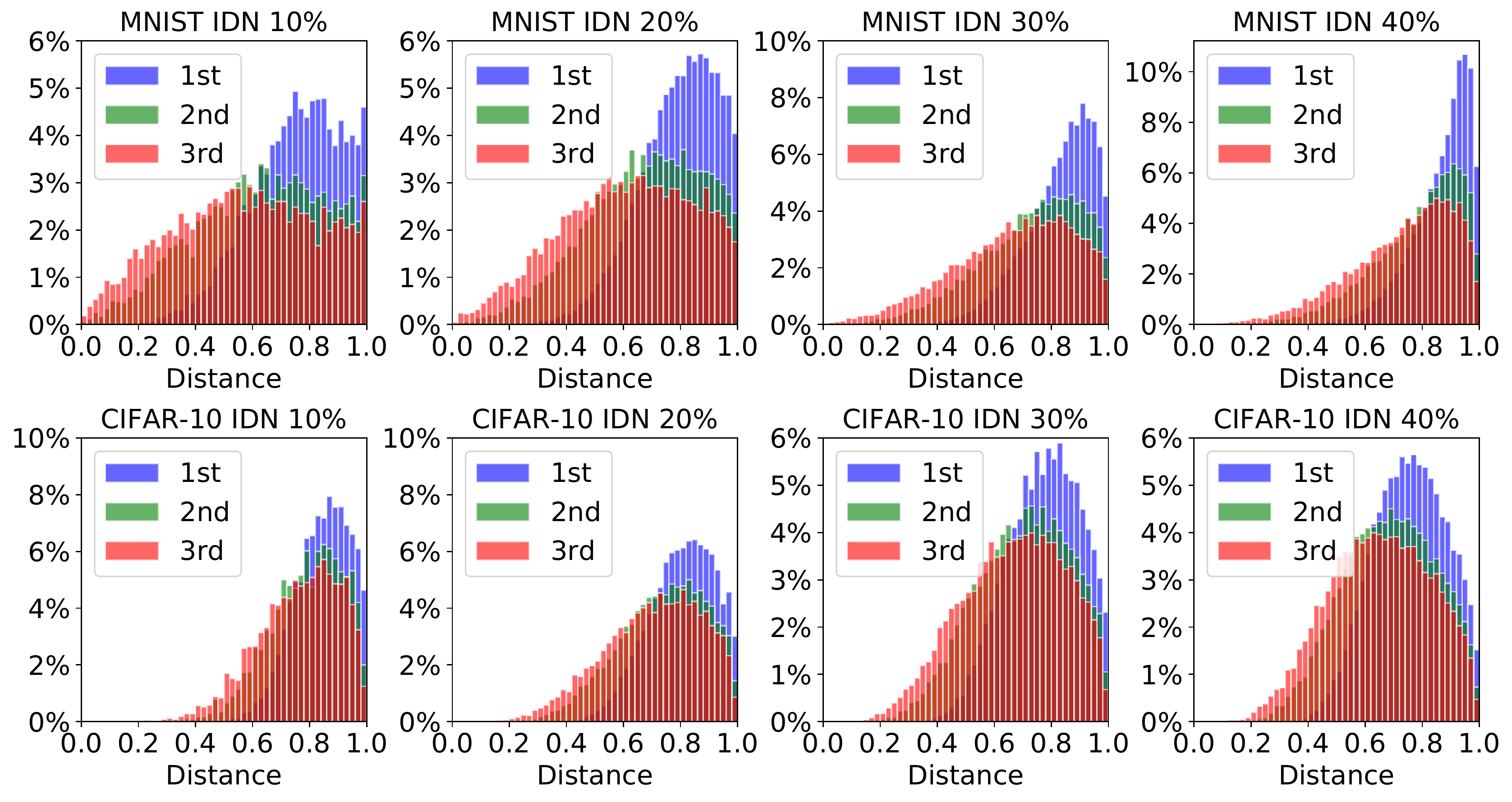}}
		\vskip -0.15in
		\caption{Histograms of distance distribution, where distance is evaluated between the true label and the soft label obtained by SEAL in 1-3 iteration.}
		\label{fig_label_distance}
	\end{center}
	\vskip -0.2in
\end{figure}
\begin{figure}[!t]
	\begin{center}
		\centerline{\includegraphics[width=\columnwidth]{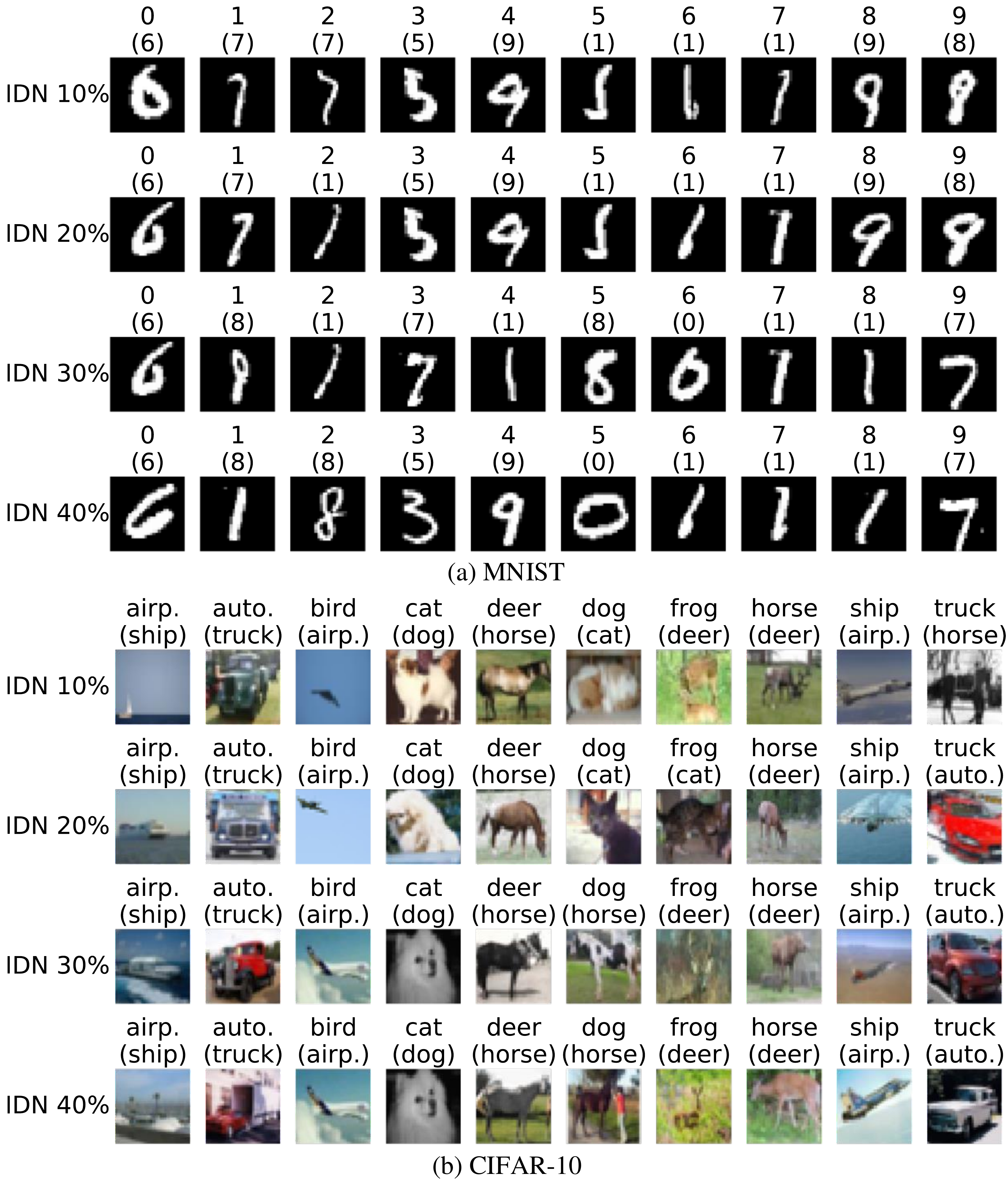}}
		\vskip -0.15in
		\caption{The noisy label and label correction (in parentheses) obtained from SEAL. The airp. and auto. are airplane and automobile for short.}
		\label{fig_correction}
	\end{center}
	\vskip -0.2in
\end{figure}

We first evaluate distances between the true label and the soft label obtained by SEAL for all noisy instances. For each noisy instance, the distance is 
\begin{equation}
d(\bar{S}_{i},y_i) = \|\bar{S}_{i}-e_{y_i}\|_1 / \|e_{\bar{y}_i}-e_{y_i}\|_1
\end{equation}
where the denominator is to normalize the distance such that $d(\bar{S}_{i},y_i)\in[0,1]$. Before running SEAL, the label is initialized as the one-hot observed label $e_{\bar{y}_i}$, hence the distance concentrates at 1.0. In Fig.~\ref{fig_label_distance}, we show histograms of distance distribution for all noisy instances. When running SEAL iteratively, the distribution moves toward the left (distance reduced), suggesting that the updated soft label approaches the true label. This verifies that SEAL can correct label noise on varying datasets and noise fractions. To further investigate individual instances, we define $\bar{N}(x)$-the confidence that a label needs correction and $\tilde{y}(x)$-the label correction
\begin{equation}
\bar{N}(x_i)=\max_{k\neq \bar{y}_i}\bar{S}_{i,k},\quad\tilde{y}(x_i)=\arg\max_{k\neq \bar{y}_i}\bar{S}_{i,k}.
\end{equation}
In Fig.~\ref{fig_correction}, we present examples with the highest $\bar{N}(x)$ in each class, with the noisy observed label and proposed label correction (in parentheses) annotated on top of each image. SEAL can identify and correct noisy labels.

\paragraph{SEAL improves generalization under IDN.}

\begin{table}[!t]
	\caption{Classification accuracies ($\%$) on MNIST under instance-dependent label noise with different noise fractions.}
	\label{table_mnist_idn}
	\vskip -0.3in
	\begin{center}
		\begin{small}
			\begin{tabular}{lrrrr}
				\toprule[1pt]
				Method	 &10$\%$ &20$\%$ &30$\%$ &40$\%$\\
				\midrule[1pt]
				\multirow{2}{*}{CE}			&94.07		&85.62		&75.75		&65.83 \\
				&$\pm$0.29  &$\pm$0.56  &$\pm$0.09  &$\pm$0.56\\
				\midrule							
				\multirow{2}{*}{Forward}	&93.93		&85.39		&76.29		&68.30 \\
				&$\pm$0.14  &$\pm$0.92  &$\pm$0.81  &$\pm$0.42\\
				\midrule							
				\multirow{2}{*}{Co-teaching}&95.77		&91.07		&86.20		&79.30 \\
				&$\pm$0.03  &$\pm$0.19  &$\pm$0.35  &$\pm$0.84\\
				\midrule
				\multirow{2}{*}{GCE}			&94.56		&86.71		&78.32		&69.78 \\
				&$\pm$0.31  &$\pm$0.47  &$\pm$0.43  &$\pm$0.58\\
				\midrule						
				\multirow{2}{*}{DAC}		&94.13		&85.63		&75.82		&65.69 \\
				&$\pm$0.02  &$\pm$0.56  &$\pm$0.58  &$\pm$0.78\\
				\midrule							
				\multirow{2}{*}{DMI}	&94.21		&87.02		&76.19		&67.65 \\
				&$\pm$0.12  &$\pm$0.42  &$\pm$0.64  &$\pm$0.73\\
				\midrule							
				\multirow{2}{*}{SEAL} &\textbf{96.75}		&\textbf{93.63}		&\textbf{88.52}		&\textbf{80.73} \\
				&$\pm$0.08  &$\pm$0.33  &$\pm$0.15  &$\pm$0.41\\				
				\bottomrule[1pt]
			\end{tabular}
		\end{small}
	\end{center}
\end{table}
\begin{table}[!t]
	\caption{Classification accuracies ($\%$) on CIFAR-10 under instance-dependent label noise with different noise fractions.}
	\label{table_cifar10_idn}
	\vskip -0.3in
	\begin{center}
		\begin{small}
			\begin{tabular}{lrrrr}
				\toprule[1pt]
				Method	 &10$\%$ &20$\%$ &30$\%$ &40$\%$\\
				\midrule[1pt]
				\multirow{2}{*}{CE}			&91.25		&86.34		&80.87		&75.68 \\
				&$\pm$0.27  &$\pm$0.11  &$\pm$0.05  &$\pm$0.29\\
				\midrule							
				\multirow{2}{*}{Forward}	&91.06		&86.35		&78.87		&71.12 \\
				&$\pm$0.02  &$\pm$0.11  &$\pm$2.66  &$\pm$0.47\\
				\midrule							
				\multirow{2}{*}{Co-teaching}&91.22		&87.28		&84.33		&78.72 \\
				&$\pm$0.25  &$\pm$0.20  &$\pm$0.17  &$\pm$0.47\\
				\midrule
				\multirow{2}{*}{GCE}		&90.97		&86.44		&81.54		&76.71 \\
				&$\pm$0.21  &$\pm$0.23  &$\pm$0.15  &$\pm$0.39\\
				\midrule							
				\multirow{2}{*}{DAC}		&90.94		&86.16		&80.88		&74.80 \\
				&$\pm$0.09  &$\pm$0.13  &$\pm$0.46  &$\pm$0.32\\
				\midrule							
				\multirow{2}{*}{DMI}		&91.26		&86.57		&81.98		&77.81 \\
				&$\pm$0.06  &$\pm$0.16  &$\pm$0.57  &$\pm$0.85\\
				\midrule							
				\multirow{2}{*}{SEAL}	&\textbf{91.32}		&\textbf{87.79}		&\textbf{85.30}		&\textbf{82.98} \\
				&$\pm$0.14  &$\pm$0.09  &$\pm$0.01  &$\pm$0.05\\				
				\bottomrule[1pt]
			\end{tabular}
		\end{small}
	\end{center}
\end{table}

We conduct experiments on MNIST and CIFAR-10 with IDN of varying noise fractions, compared with extensive baselines including (i) cross-entropy (CE) loss; (ii) Forward~\cite{patrini2017making}, which trains a network to estimate an instance-\textit{in}dependent noise transition matrix then corrects the loss; (iii) Co-teaching~\cite{han2018co}, where two classifiers select small-loss instances to train each other; (iv) Generalized Cross Entropy (GCE) loss, which is a robust version of CE loss with theoretical guarantee under CCN; (v) deep abstaining classifier (DAC)~\cite{thulasidasan2019combating}, which gives option to abstain samples depending on the cross-entropy error and an abstention penalty; (vi) Determinant based Mutual Information (DMI), which is an information-theoretic robust loss. The number of iterations is $10$ on MNIST and $3$ on CIFAR-10. SEAL consistently achieves the best generalization performance, as shown in Table~\ref{table_mnist_idn} and Table~\ref{table_cifar10_idn}, where we report the accuracy at the last epoch and repeat each experiment three times.

\paragraph{SEAL improves generalization under real-world noise.}

\begin{table}[t]
	\caption{Testing accuracy ($\%$) on Clothing1M. The $*$ marks published results.}
	\label{table_clothing}
	\vskip -0.3in
	\begin{center}
		\begin{small}
			\begin{tabular}{lcc}
				\toprule[1pt]
				Method					&Accuracy \\
				\midrule[1pt]
				CE$*$	                    &68.94 \\
				Forward$*$	                &69.84 \\
				Co-teaching			        &70.15 \\
				GCE$*$	                    &69.09 \\
				Joint Optimization$*$		&72.16 \\
				DMI$*$				        &72.46 \\
				\midrule
				CE					        &69.07 \\
				SEAL			            &\textbf{70.63} \\
				\midrule
				DMI					        &72.27 \\
				SEAL (DMI)		            &\textbf{73.40} \\			
				\bottomrule[1pt]
			\end{tabular}
		\end{small}
	\end{center}
\end{table}

Clothing1M~\cite{xiao2015learning} is a large-scale real-world dataset of clothes collected from shopping websites, with noisy labels assigned by the surrounding text. Following the benchmark setting~\cite{patrini2017making,tanaka2018joint,xu2019l_dmi}, the training set consists of $1M$ noisy instances and the additional validation, testing sets consist of $14K$, $10K$ clean instances. The number of SEAL iterations is $3$. In Table~\ref{table_clothing}, we present the test accuracy. By default, SEAL is implemented with normal cross-entropy, where we see $1.56\%$ absolute improvement. Notably, SEAL also improves advanced training algorithms such as DMI~\cite{xu2019l_dmi} when we use the method as initialization.

\section{Conclusion}
In this paper, we theoretically justify the urgent need to go beyond the CCN assumption and study IDN. We formalize an algorithm to generate controllable IDN which is semantically meaningful and challenging. As a primary attempt to combat IDN, we propose a method SEAL, which is effective for both synthetic IDN and real-world noise.

Notably, our theoretical analysis in Section~\ref{sec_ccn2idn} provides rigorous motivations for studying IDN. Learning with IDN is an important topic that deserves more research attention in future.

\clearpage

\section{Acknowledgments}
The work is supported by the Key-Area Research and Development Program of Guangdong Province, China (2020B010165004) and the National Natural Science Foundation of China (Grant Nos.: 62006219, U1813204).

\bibliography{paper}

\clearpage
\onecolumn
\appendix

\section{Examples of noisy samples}
\label{app_eg_idn}

\begin{figure*}[ht]
	\begin{center}
		\centerline{\includegraphics[width=0.9\columnwidth]{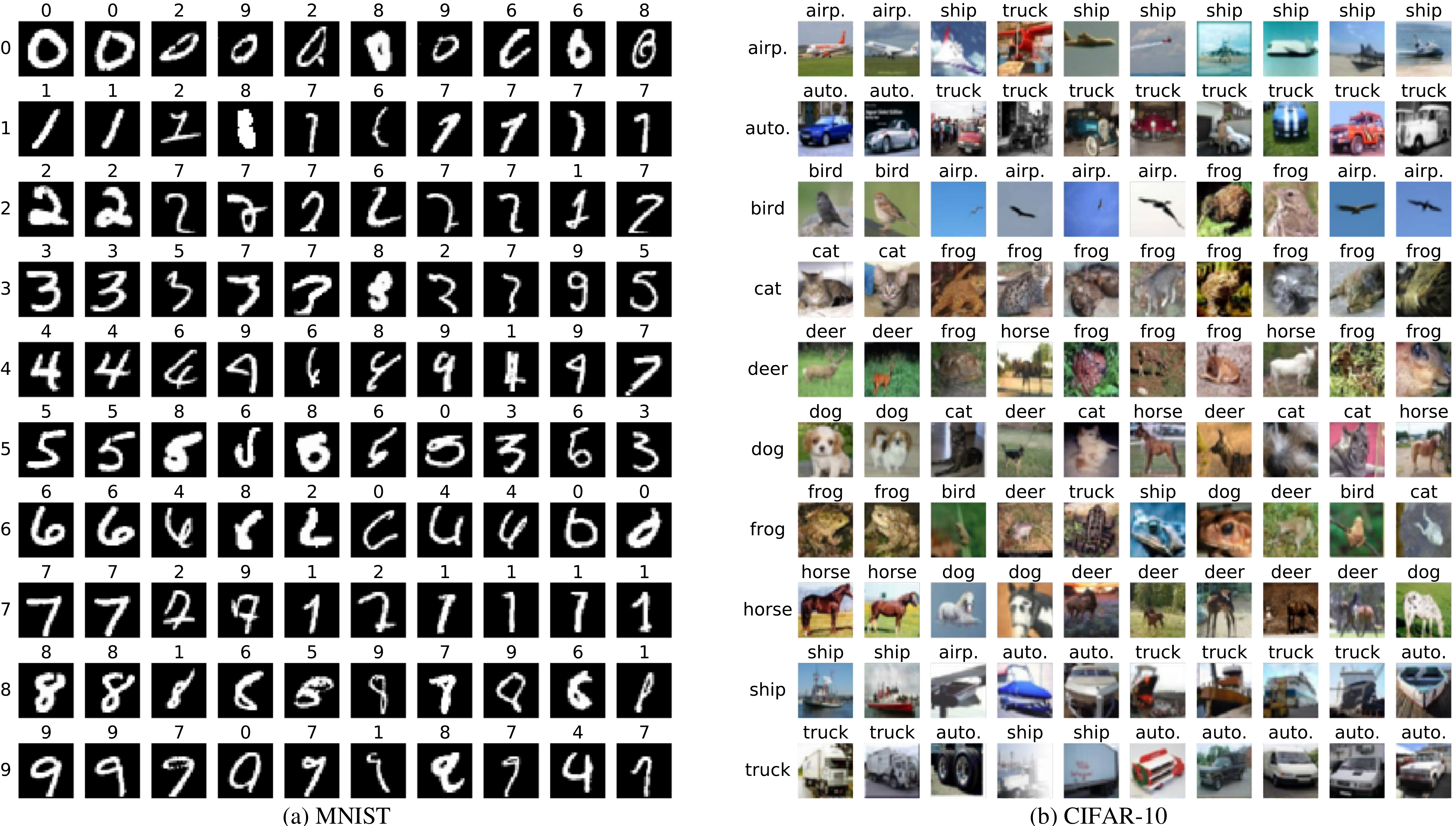}}
		\caption{Our IDN labeler generates label noise (on top of each image) highly dependent on instances. As an example, each row corresponds to instances of the same true class, consisting of two correct instances and eight mislabeled instances. The airp. and auto. are airplane and automobile for short.}
		\label{fig_dependent}
	\end{center}
\end{figure*}

\section{More details on experiments}
\label{app_exp_setup}

\begin{itemize}

\item  On \textbf{MNIST}, we use a convolution neural network (CNN) with the standard input 28$\times$28 and 4 layers as follows: [conv 5$\times$5, filters 20, stride 1, relu, maxpool /2]; [conv 5$\times$5, filters 50, stride 1, relu, maxpool /2]; [fully connect 4*4*50$\rightarrow$500, relu]; [fully connect 500$\rightarrow$10, softmax]. Models are trained for 50 epochs with a batch size of 64 and we report the testing accuracy at the last epoch. For the optimizer, we use SGD with a momentum of 0.5, a learning rete of 0.01, without weight decay.

\item  On \textbf{CIFAR-10}, we use the Wide ResNet 28$\times$10. Models are trained for 150 epochs with a batch size of 128 and we report the testing accuracy at the last epoch. From Fig~\ref{fig_train_curve} in the main paper, we can see that the epoch of $150$ is sufficient large for the training accuracy to converges to $100\%$. For the optimizer, we use SGD with a momentum of 0.9 and a weight decay of $5\times10^{-4}$ The learning rate is initialized as 0.1 and is divided by 5 after 60 and 120 epochs. We apply the standard data augmentation on CIFAR-10:  horizontal random flip and 32$\times$32 random crop after padding 4 pixels around images. The standard normalization with mean=(0.4914, 0.4822, 0.4465), std=(0.2023, 0.1994, 0.2010) is applied before feeding images to the network.

\item  On \textbf{Clothing1M}, following the benchmark setting~\cite{patrini2017making,tanaka2018joint,xu2019l_dmi}, we use the ResNet-50 pre-trained on ImageNet and access the clean validation set consisting of $14K$ instances to do model selection. Models are trained for 10 epochs with a batch size of 256 on the noisy training set consisting of $1M$ instances. For the optimizer, we use SGD with a momentum of 0.9 and a weight decay of $10^{-3}$. We use a learning rate of $10^{-3}$ in the first 5 epochs and $10^{-4}$ in the second 5 epochs in all experiments except for DMI~\cite{xu2019l_dmi}, where the learning rate is $10^{-6}$ and $0.5\times10^{-6}$ according to its original paper. We apply the standard data augmentation: horizontal random flip and 224$\times$224 random crop. Before feeding images to the network, we normalize each image with mean and std from ImageNet, i.e., mean=(0.485, 0.456, 0.406), std=(0.229, 0.224, 0.225). Considering that a pre-trained model and a clean validation are accessed in all methods, we do not reinitialize our model in each SEAL iteration, instead, we start the training on top of the best model from the last iteration.

\end{itemize}

\end{document}